# Silicon Minds versus Human Hearts: The Wisdom of Crowds Beats the Wisdom of AI in Emotion Recognition

**Mustafa Akben**[1] Elon University, NC, USA
**Vinayaka Gude** Elon University, NC, USA
**Haya Ajjan** Elon University, NC, USA

## Abstract

*The ability to discern subtle emotional cues is fundamental to human social intelligence. As artificial intelligence (AI) becomes increasingly common, AI's ability to recognize and respond to human emotions is crucial for effective human-AI interactions. In particular, whether such systems can match or surpass human experts remains to be seen. However, the emotional intelligence of AI, particularly multimodal large language models (MLLMs), remains largely unexplored. This study evaluates the emotion recognition abilities of MLLMs using the Reading the Mind in the Eyes Test (RMET) and its multiracial counterpart (MRMET), and compares their performance against human participants. Results show that, on average, MLLMs outperform humans in accurately identifying emotions across both tests. This trend persists even when comparing performance across low, medium, and expert-level performing groups. Yet when we aggregate independent human decisions to simulate collective intelligence, human groups significantly surpass the performance of aggregated MLLM predictions, highlighting the wisdom of the crowd. Moreover, a collaborative approach (augmented intelligence) that combines human and MLLM predictions achieves greater accuracy than either humans or MLLMs alone. These results suggest that while MLLMs exhibit strong emotion recognition at the individual level, the collective intelligence of humans and the synergistic potential of human-AI collaboration offer the most promising path toward effective emotional AI. We discuss the implications of these findings for the development of emotionally intelligent AI systems and future research directions.*

## 1. Introduction

Emotion recognition, which refers to the ability to discern and interpret others' emotional states, influences how individuals perceive, think, and navigate their social environments (Salovey and Mayer, 1990). People adept at recognizing emotions forge deep connections with others, resolve conflicts, and sustain social cohesion (Lopes et al., 2004; Vellante et al., 2013), whereas deficits present difficulties ranging from neuropsychiatric disorders to socio-cognitive challenges (Paiva-Silva et al., 2016; Vellante et al., 2013).

This same capacity has become a sine qua non for artificial intelligence (AI) systems as they become increasingly ubiquitous in daily life (Picard, 2010). AI systems are currently entrusted with high-stakes emotion recognition tasks that require near-perfection, from identifying depression in psychiatric clinics to detecting deception at border security (Garcia-Ceja et al., 2018; Sánchez-Monedero and Dencik, 2022). In such conditions, emotion recognition is not an optional feature but a fundamental requirement to inspire trust, encourage cooperation, and ultimately deliver success. Against this backdrop, researchers have accelerated progress on emotion-aware AI systems, yielding substantial advances (Monedero and Dencik, 2022).

Leading this progress, large language models (LLMs) have advanced rapidly. Over the past few years, LLMs have now performed at average human levels or better on widely accepted text-based emotion detection tests, as documented in recent studies (Elyoseph et al., 2023; Wang et al., 2023). More recently, multimodal LLMs (MLLMs) capable of processing both language and images have outperformed average individuals on image-based tests (Strachan, Pansardi, et al., 2024). GPT-4o from OpenAI, consistently ranked highest among multimodal models in benchmarks and emotion recognition tests (Hurst et al., 2024), exemplifies current state-of-the-art performance. Accordingly, GPT-4o serves as an ideal test case for evaluating current AI capabilities in emotion recognition. However, prevailing assessment methods may not capture the model's full range of abilities.

[1] Corresponding author: makben@elon.edu

Specifically, conventional methodologies tend to rely on central tendency metrics (e.g., group means) to compare AI model performance to the average human level. Illustratively, Strachan et al. (2024) found that GPT-4o exceeded average human performance, but their analysis relied solely on group mean differences, thereby failing to test whether AI could match the performance of expert-level humans. Such approaches overlook the wide variation in human emotion recognition abilities and the exceptional performers whose skills are critical in high-stakes fields like clinical psychology and medical practice (Ickes, 2009). Validating GPT-4o only against average performance thus risks deploying suboptimal systems in high-stakes settings and mistaking AI mediocrity for human excellence. To address this gap, we pose our first research question (RQ1): How does GPT-4o's emotion recognition performance compare across low, medium, and top-performing human groups, rather than only against the average?

Our second research question (RQ2) extends the first by examining whether GPT-4o can outperform collective human intelligence, or the "wisdom of crowds." This comparison is critical because society's important decisions often emerge from groups rather than individuals (Centola, 2018; Surowiecki, 2005), with collective judgment canceling individual biases to achieve better accuracy (Hong & Page, 2004). Whether AI can match this collective intelligence thus provides a litmus test of its true capabilities, especially given claims of human-level or better emotional skills (Elyoseph et al., 2023; Mittelstädt et al., 2024; Refoua, Meinlschmidt, and Elyoseph, 2024).

To test RQ2, we implemented a plurality voting procedure for both human and GPT-4o conditions. This plurality voting involved creating groups of varying sizes (from n = 5 to n = 90) by randomly sampling participants from the human pool and independent GPT-4o instances from our multiple runs. We then examined each selected participant's or instance's independent response to determine the group's collective decision through plurality voting. Specifically, within each group, the emotion category that received the most votes became the collective decision of that group. For example, if 7 out of 10 respondents in a group (whether human or GPT-4o) selected "jealous" for a particular item, "jealous" was designated as that group's response. We repeated this process multiple times using random sampling, generating human and AI group decisions with different group sizes (see the Methods section for details). This aggregation procedure enables a direct comparison of human and AI crowds, revealing whether GPT-4o can reproduce the statistical processes that underpin the wisdom of the crowd.

Finally, we investigate augmented intelligence by asking (RQ3) whether combining human and GPT-4o collective judgments yields superior performance to either crowd alone. This approach tests whether the complementary strengths of AI's computational abilities and human cognitive diversity can synergize to improve emotion recognition accuracy (Dellermann et al., 2019), moving beyond simple comparisons to explore genuine human-AI collaboration.

Although valuable in some settings, direct human-machine interactions can suffer from human biases, such as algorithm aversion or overreliance on AI inputs, which may reduce rather than enhance collaborative performance (Bansal et al., 2021; Prahl and Van Swol, 2017). To circumvent these interaction biases, we aggregate independent human and AI predictions rather than having them interact directly, allowing each to contribute strengths without interference. We used a similar majority voting system as described previously. This independent aggregation design, particularly the use of plurality voting to create augmented intelligence, represents a novel approach to emotion recognition tasks that remains largely unexplored.

Taken together, our study investigates three questions about GPT-4o's emotion recognition capabilities. (RQ1) Does GPT-4o consistently outperform humans at all skill levels, or only at specific points in the performance distribution? (RQ2) When both human and GPT-4o predictions are aggregated through plurality voting, which collective intelligence performs better? (RQ3) Can we achieve superior emotion recognition by combining human and GPT-4o crowd predictions into a unified augmented intelligence system?

To answer these questions, we follow prior work (Strachan, Pansardi, et al., 2024) that evaluates the emotion recognition skill of humans and LLMs with the Reading the Mind in the Eyes Test (RMET) (Baron-Cohen et al., 2001) and the Multiracial Reading the Mind in the Eyes Test (MRMET) (Kim et al., 2024).

For our analysis, we used publicly available human performance data from Kim et al. (2024) (N = 17,680 for RMET and N = 9,295 for MRMET) and generated GPT-4o responses via the API. We evaluated accuracy using mixed-effects logistic regression, shift function analyses, and stochastic dominance analyses, thereby capturing performance across the full range of abilities. To test augmented intelligence, we aggregated GPT-4o and human judgments via plurality voting and assessed whether the combined approach outperformed either source alone. Our research thus provides a test of GPT-4o's true emotion-recognition capabilities and the viability of AI-assisted collaboration.

## 2. Methods

### 2.1. Participants and Procedures

In this study, we used publicly available data from Kim et al. (2024), who collected 17,680 and 9,295 responses for the RMET and MRMET, respectively, and described original demographic information, recruitment methods, and consent information. For our study, we ran GPT-4o in February 2025 for 50 trials per condition (N = 50 for zero-shot; N = 50 for 10-shot) with default settings. Following machine learning evaluation practices (Brown et al., 2020), we assessed the model under two prompting conditions: (a) zero-shot, to evaluate its baseline performance without examples, and (b) 10-shot, to test its optimal performance with them.

In the zero-shot condition, the model analyzed each image from the RMET and MRMET datasets. Each image included the instruction: "Please pick the best possible answer that describes the emotion or mental state the person is experiencing from the following options." This zero-shot condition provided no examples or demonstrations, testing the model's baseline ability to infer emotion from pre-trained knowledge and generalization.

In contrast, the 10-shot condition evaluated the model's capacity for few-shot contextual learning. The model received ten examples of randomly selected images from the dataset, and each was paired with its corresponding ground-truth answer. This 10-shot approach investigated the model's ability to learn from a limited number of examples. We used 10-shot examples based on prior research showing that performance gains in few-shot settings often plateau after a small number of examples (Brown et al., 2020; Wei et al., 2022), and that too many examples can adversely affect the model's performance (Min et al., 2022).

As detailed in the Findings section, we later found no significant differences between the zero-shot and 10-shot conditions, so we combined these conditions in subsequent analyses. This pooling decision simplified interpretation and provided a comprehensive estimate of GPT-4o's performance, capturing the model's abilities from baseline to enhanced performance.

We ran the model with its default settings: a temperature of 1 and top-p (nucleus) sampling of 1. Sample size differences between human participants (17,680 for RMET and 9,295 for MRMET) and GPT-4o runs (50 for zero-shot and 50 for 10-shot) were addressed through statistical approaches. LLMs like GPT-4o have low variance between runs, showing consistency even with high temperature and top-p sampling (Chang et al., 2023), especially with multiple-choice options. Previous work (Chen et al., 2024) shows statistical performance gains for LLMs diminish sharply after a few runs, making large samples unnecessary. To account for the uneven sample sizes, we used mixed-effects models that adjust through partial pooling, allowing flexible group-size handling (Pinheiro and Bates, 2000). Thus, these sample size differences between conditions do not threaten our findings' validity or reliability.

### 2.2. Materials

To assess emotion recognition abilities, we employed two widely used and validated instruments. The Reading the Mind in the Eyes Test (RMET) is a validated test for inferring mental states and emotions (Baron-Cohen et al., 2001). It uses 36 black-and-white photographs of the eye region. Participants select from four options the label best describing the person's mental state. Applied across diverse populations, the RMET demonstrates adequate psychometric properties.

The Multiracial Reading the Mind in the Eyes Test (MRMET), a recently validated test (Kim et al., 2024), addresses RMET weaknesses, such as a lack of racial diversity and inaccessible vocabulary. It uses 37 high-resolution, color photographs with diverse participants. The MRMET demonstrates sufficient psychometric properties for emotion recognition tasks. For the validation studies, please refer to Kim et al., 2024.

Finally, all statistical analyses were conducted using R version 4.3.2. GPT-4o data collection used Python 3.9 via the OpenAI API. Code and data are available upon request.

### 2.3. Analytical strategy

***Logistic Mixed-Effects Models.*** To analyze these data and address our research questions, we employed four complementary statistical approaches. First, we used generalized linear mixed-effects models (GLMMs) with a logit link, as outcomes were binary (correct = 1 and incorrect = 0) (Snijders and Bosker, 2012). Our data had a nested structure. Test items were nested within participants, who were themselves nested within groups. GLMMs suit this structure, accounting for random effects at item and group levels. These models handle unbalanced designs, especially with 30+ observations per group (Meteyard and Davies, 2020).

We used random intercepts to model test items and participants, and fixed effects for group conditions (e.g., human, GPT-4o zero-shot, GPT-4o 10-shot). We used the human condition as the reference group. Given dataset size and potential convergence problems, we implemented maximum likelihood estimation with adaptive Gauss-Hermite quadrature. Models were fitted using the lme4 package with diagnostics for overdispersion and convergence. Model fit used likelihood ratio tests with ORs and 95% confidence intervals. Odds ratios were derived using the emmeans package. Odds ratios greater than 1 indicated higher accuracy than the reference human group. *p*-values were adjusted using Bonferroni corrections.

***Bootstrap Shift-Function Test for Quantile Performance.*** Beyond mean comparisons, to analyze performance across quantiles, we used a bootstrap shift-function analysis with the "rogme" R package (Rousselet, 2019). This shift function shows differences between two distributions by comparing them at multiple quantiles, examining how one distribution shifts and changes across the other. We compared human and GPT-4o performance across quantiles ranging from 0.03 to 0.97, excluding extremes (0.01 and 0.99), where accuracy saturates at 0 or 1 due to ceiling effects. Given our finding of no significant differences between the zero- and 10-shot conditions through our GLMMs, we combined them to create a single GPT-4o condition. This pooling efficiently assesses the model's full range, from baseline performance to enhanced contextual learning.

Performance scores were calculated by averaging item accuracy per participant for humans and per trial for GPT-4o. For example, if someone predicts 35 of 36 RMET items correctly, accuracy equals 35/36 = 0.972. We used the Harrell-Davis quantile estimator. To assess between-group differences across distributions, we computed uncertainty intervals for each quantile comparison using percentile bootstrap techniques. This approach determined whether performance differences between humans and GPT-4o varied across performance segments, providing uncertainty estimates without parametric assumptions (Wilcox, 2011). We used Bonferroni corrections for multiple comparisons. Results were plotted as shift functions for each comparison, showing distributional differences across quantiles. Analyses were carried out independently for RMET and MRMET tasks.

***First and Second-Order Stochastic Dominance Tests.*** In addition to shift-function analysis, we conducted first- and second-order stochastic dominance analyses to assess distributions. These analyses compare entire performance spectra to determine whether one group systematically outperforms another at every accuracy threshold. Stochastic dominance analyses involve comparisons of entire distributions rather than sample characteristics like means or quantiles(Davidson and Duclos 2000); for comprehensive reviews, see(Whang 2019). We used aggregated performance scores per participant and compared group conditions (human versus GPT-4o) as before.

More specifically, first-order stochastic dominance (FSD) exists when one distribution consistently offers higher probabilities of achieving values greater than or equal to any given level compared to another distribution. Formally, for cumulative distribution functions $F_A(x)$ and $F_B(x)$ of distributions $A$ and $B$, $A$ exhibits FSD over $B$ if $F_A(x) \leq F_B(x)$ for all $x$, with strict inequality for at least one $x$. This implies that, at any performance threshold, the dominant distribution has a higher probability of surpassing that threshold. Second-order stochastic dominance (SSD) compares the overall shapes of two distributions, considering both their central tendencies and their spread (i.e., risk or variability). It extends FSD. Unlike FSD, which requires one cumulative distribution function, or CDF, to lie entirely below the other, SSD allows the CDFs to cross once or twice if the area under one remains smaller overall. Distribution A exhibits SSD over $B$ if $\int_{-\infty}^{x} F_A(t)\,dt \leq \int_{-\infty}^{x} F_B(t)\,dt$ for all $x$, with strict inequality for at least one $x$. If $A$ shows SSD, it suggests more consistent and stable performance across various test items.

We used bootstrap with 2,000 replicates to calculate CDFs for both groups across all accuracy levels(Wilcox 2011). We tested dominance in both directions (AI over humans and humans over AI) where the null hypothesis posits that dominance exists. Critically, this reverses typical statistical interpretation: p-values > 0.05 support dominance by failing to reject the null, while p-values < 0.05 indicate dominance violations. Readers should note this inverted interpretation when evaluating our results. If AI demonstrated both FSD and SSD over the human condition, this indicates consistent AI superiority. Conversely, partial overlap or absent stochastic dominance suggests advantages only at specific segments, revealing performance consistency between conditions. Analyses were carried out for the RMET and MRMET tasks.

***Plurality Voting: Artificial, Collective, and Augmented Intelligence.*** Finally, to address our questions about collective and augmented intelligence, we analyzed accuracy at the group level for human groups, GPT-4o groups, and combined human-AI groups using a plurality voting procedure. This procedure creates "crowds" by aggregating individual responses at the group level, where the emotion category receiving the most votes becomes the collective decision. For human crowds, we randomly sampled participants from the total pool to form groups of varying sizes. We created groups containing 5 to 90 participants, with replacement across groups. For each test item, we counted how many individuals in the sampled group selected each emotion option, and the option with the most votes became that crowd's answer. For GPT-4o crowds, we applied the identical plurality voting method, collecting the model's responses from multiple independent runs and forming groups of the same size as the human condition. This parallel approach ensures that both human and AI crowds are formed through the same aggregation process.

To create the augmented intelligence (human + AI) crowds, we combined human and GPT-4o responses using the same plurality voting procedure. Given the larger human sample size, we used a stratified approach, sampling humans and GPT-4o responses at a 10:1 ratio to form mixed crowds with 90% human and 10% AI participants. For each test item, we counted votes from both human and AI participants together, and the emotion receiving the most votes became the augmented crowd's answer. In the case of a tie, we randomly selected one of the tied choices.

To compare group performance, we used GLMMs with a logit link as described above. These models included random intercepts for both test items and iteration, and fixed effects for group condition (human, GPT-4o, human + GPT-4o) and log-transformed group size. Log transformation accounted for diminishing returns in collective decisions (Navajas et al., 2018; Peeters et al., 2021). Separate analyses were performed for RMET and MRMET datasets, testing main effects and interactions. Marginal slopes used the emtrends function in emmeans, with interactions visualized via the interactions R package.

# 3. Results

## 3.1. AI Outperformed Average Human Performance on Emotion Recognition Tasks

To address our first research question about performance across ability levels, we begin by examining overall performance differences between humans and GPT-4o. We evaluated human and GPT-4o performance differences using RMET and MRMET tasks. Descriptive statistics and average scores are presented in Table 1. On the RMET, GPT-4o achieved 0.90 (SD = 0.03) (pooled across zero-shot and 10-shot conditions), outperforming the human average of 0.71 (SD = 0.14): 0.89 (zero-shot) and 0.91 (10-shot). On the MRMET, GPT-4o attained 0.83 (SD = 0.03) (pooled), surpassing the human average of 0.62 (SD = 0.13): 0.81 (zero-shot) and 0.85 (10-shot). While GPT-4o improved slightly with examples (10-shot condition), these differences were not statistically significant (see Supplementary Table S1–S2). Therefore, we combined zero-shot and 10-shot conditions in subsequent analyses, simplifying interpretation without losing significant information.

**Table 1. Summary Statistics by Conditions and Tasks**

| Dataset | Model | Mean Accuracy | Standard Deviation | Max Accuracy | Min Accuracy | Sample Size (n) |
|---|---|---|---|---|---|---|
| RMET | Human | 0.71 | 0.14 | 1.00 | 0.11 | 17,680 |
| | GPT-4o (Pooled) | 0.90 | 0.03 | 0.97 | 0.84 | 100 |
| | GPT-4o (Zero-shot) | 0.89 | 0.03 | 0.95 | 0.84 | 50 |
| | GPT-4o (10-shot) | 0.91 | 0.03 | 0.97 | 0.86 | 50 |
| MRMET | Human | 0.62 | 0.13 | 0.97 | 0.03 | 9,295 |
| | GPT-4o (Pooled) | 0.83 | 0.03 | 0.92 | 0.74 | 100 |
| | GPT-4o (Zero-shot) | 0.81 | 0.02 | 0.84 | 0.76 | 50 |
| | GPT-4o (10-shot) | 0.85 | 0.03 | 0.92 | 0.74 | 50 |

**Notes.** RMET = Reading the Mind in the Eyes Test; MRMET = Multiracial Reading the Mind in the Eyes Test. Accuracy is calculated as the proportion of correct responses across all items. Mean Accuracy represents the average performance across all trials for each condition. Standard Deviation indicates the variability in performance within each condition. Max and Min Accuracy represent the highest and lowest performance achieved by any single participant/iteration within each condition. Sample Size (*n*) for human participants represents individual participants, while for GPT-4o conditions, it represents the number of independent model iterations. Human participants completed all items in each test. GPT-4o models were tested in both zero-shot (no examples

provided) and 10-shot (10 examples provided) conditions. GPT-4o (Pooled) combines both zero-shot and 10-shot iterations ($n$ = 100 total). Human data remains the same whether analyzed separately or pooled, as all participants completed the same task.

With the combined pooled results, we ran the rest of the analysis. Results from generalized linear mixed-effects models (GLMMs) show the observed differences were significant. Overall, GPT-4o achieved higher accuracy than humans on both RMET and MRMET, consistent with prior findings. Table 2 shows the results for the pooled GPT-4o analysis. On the RMET, GPT-4o showed significantly higher log-odds than humans ($\beta = 1.27$, SE = 0.08, $p < 0.001$). On the MRMET, GPT-4o showed significantly higher log-odds than humans ($\beta = 1.11$, SE = 0.07, $p < 0.001$). For separate zero-shot and 10-shot results, see Supplementary Table S3.

As such, the results show that, on average, GPT-4o was much more accurate than average human beings in both RMET and MRMET tests. However, average scores can obscure important differences in performance distributions across skill levels and are less likely to reveal whether GPT-4o can outperform the best-performing human groups. In our second analysis, we explore these questions.

**Table 2. Mixed-Effects Logistic Regression Results for RMET and MRMET**

| Parameter | RMET | | | MRMET | | |
|---|---|---|---|---|---|---|
| | Estimate | SE | p-value | Estimate | SE | p-value |
| Fixed Effects | | | | | | |
| (Intercept) | 0.96 | 0.08 | p < 0.001 | 0.69 | 0.15 | p < 0.001 |
| GPT-4o | 1.27 | 0.08 | p < 0.001 | 1.11 | 0.07 | p < 0.001 |
| Random Effects | | | | | | |
| Participant Variance | 0.39 | - | - | 0.34 | - | - |
| Participant Std Dev | 0.62 | - | - | 0.59 | - | - |
| Item Variance | 0.21 | - | - | 0.88 | - | - |
| Item Std Dev | 0.46 | - | - | 0.94 | - | - |
| Model Information | | | | | | |
| Observations | 640,080 | - | - | 347,615 | - | - |
| Number of Participants | 17,780 | - | - | 9,395 | - | - |
| Number of Items | 36 | - | - | 37 | - | - |
| AIC | 729,232.50 | - | - | 395,826.00 | - | - |
| BIC | 729,277.90 | - | - | 395,869.00 | - | - |
| Log Likelihood | -364,612.20 | - | - | -197,909.00 | - | - |

**Notes.** RMET = Reading the Mind in the Eyes Test; MRMET = Multiracial Reading the Mind in the Eyes Test. Both models were fit as generalized linear mixed models (GLMM) using the lme4 package in R, with a logit link function. Models were fit by maximum likelihood using adaptive Gauss-Hermite quadrature (nAGQ = 0) with crossed random intercepts for participant and item. Fixed effects estimates are presented on the log-odds scale; exponentiating these values yields odds ratios. *SE* refers to standard error. *P*-values were calculated using Satterthwaite's method for denominator degrees of freedom approximation through the lmerTest package, with $\alpha = 0.05$. Human performance serves as the reference group for all comparisons (Intercept). The GPT-4o (Pooled) parameter represents the pooled effect of both zero-shot and 10-shot conditions, which was statistically justified through model comparison (see Table 6). Random effects are reported as variances and their standard deviations, representing the unexplained variation at the participant and item levels. AIC (Akaike Information Criterion) and BIC (Bayesian Information Criterion) provide measures of model fit, with smaller values indicating better fit. Log Likelihood represents the logarithm of the model's likelihood function at convergence. Both models showed successful convergence with no warnings or errors

## 3.2 AI Outperform Human Performance Across Low-, Medium-, and High-Performing Group Levels

Moving beyond averages to address our first research question more fully, we examined performance differences across ability levels. We conducted a shift-function analysis with bootstrap resampling, comparing AI and human performance distributions across quantiles. Figure 1 shows these differences across the quantiles.

Table 3 presents quantile-specific comparisons between GPT-4o and humans, employing the Harrell-Davis quantile estimator and percentile bootstrap. Reported differences represent Δ GPT-4o – Human quantile estimates, where positive values indicate higher accuracy for GPT-4o. The analysis revealed that GPT-4o showed higher accuracy than humans across the entire performance spectrum. Notably, the magnitude of this advantage varied systematically across ability levels.

**Table 3. Quantile Comparison of Results with Shift Function**

| Dataset | Quantile | GPT-4o | Human | Difference | 95% CI Lower | 95% CI Upper | *p* | Adjusted *p* |
|---|---|---|---|---|---|---|---|---|
| **RMET** | 0.03 | 0.77 | 0.38 | 0.40 | 0.38 | 0.41 | $p < 0.001$ | $p < 0.001$ |
| | 0.05 | 0.78 | 0.42 | 0.37 | 0.35 | 0.38 | $p < 0.001$ | $p < 0.001$ |
| | 0.10 | 0.80 | 0.49 | 0.31 | 0.30 | 0.32 | $p < 0.001$ | $p < 0.001$ |
| | 0.25 | 0.82 | 0.59 | 0.22 | 0.22 | 0.24 | $p < 0.001$ | $p < 0.001$ |
| | 0.50 | 0.86 | 0.69 | 0.17 | 0.15 | 0.17 | $p < 0.001$ | $p < 0.001$ |
| | 0.75 | 0.89 | 0.78 | 0.11 | 0.11 | 0.13 | $p < 0.001$ | $p < 0.001$ |
| | 0.90 | 0.92 | 0.86 | 0.06 | 0.06 | 0.07 | $p < 0.001$ | $p < 0.001$ |
| | 0.95 | 0.93 | 0.89 | 0.04 | 0.03 | 0.06 | $p < 0.001$ | $p < 0.001$ |
| | 0.97 | 0.94 | 0.91 | 0.04 | 0.01 | 0.06 | $p < 0.001$ | $p < 0.001$ |
| **Dataset** | **Quantile** | **GPT-4o** | **Human** | **Difference** | **95% CI Lower** | **95% CI Upper** | ***p*** | **Adjusted *p*** |
| **MRET** | 0.03 | 0.76 | 0.38 | 0.38 | 0.36 | 0.41 | $p < 0.001$ | $p < 0.001$ |
| | 0.05 | 0.77 | 0.41 | 0.36 | 0.35 | 0.38 | $p < 0.001$ | $p < 0.001$ |
| | 0.10 | 0.78 | 0.46 | 0.32 | 0.31 | 0.33 | $p < 0.001$ | $p < 0.001$ |
| | 0.25 | 0.81 | 0.54 | 0.27 | 0.25 | 0.27 | $p < 0.001$ | $p < 0.001$ |
| | 0.50 | 0.82 | 0.65 | 0.17 | 0.16 | 0.19 | $p < 0.001$ | $p < 0.001$ |
| | 0.75 | 0.84 | 0.73 | 0.11 | 0.11 | 0.13 | $p < 0.001$ | $p < 0.001$ |
| | 0.90 | 0.87 | 0.80 | 0.07 | 0.04 | 0.09 | $p < 0.001$ | $p < 0.001$ |
| | 0.95 | 0.88 | 0.84 | 0.04 | 0.03 | 0.07 | $p < 0.001$ | $p < 0.001$ |
| | 0.97 | 0.90 | 0.85 | 0.05 | 0.01 | 0.07 | $p < 0.001$ | $p < 0.001$ |

**Note.** RMET = Reading the Mind in the Eyes Test. MRMET = Multiracial Reading the Mind in the Eyes Test. This table presents the shift function results comparing the GPT-4o condition to the Human condition across multiple quantiles in the RMET dataset. The shift function was generated using the Harrell-Davis quantile estimator and a percentile bootstrap approach (bootstrapped 100 times), which provides confidence intervals derived from the bootstrapped quantiles rather than from the standard errors of the difference. The Difference column reflects GPT-4o minus Human quantile estimates; positive values indicate that the GPT-4o distribution is shifted toward higher values compared to Human. Quantiles range from 0.03 to 0.97. All *p*-values were computed via percentile bootstrap and corrected for multiple comparisons using the Bonferroni method. Human serves as the reference group. The mean difference across all quantiles is 0.19.

On the RMET, accuracy differences were most pronounced at lower quantiles (Δ GPT-4o – Human = 0.40 at the 0.03 quantile), indicating that GPT-4o performs substantially better than humans with lower emotion recognition abilities (Table 3). While the difference decreases at higher quantiles, GPT-4o maintains significantly higher accuracy even at the 0.97 quantile (Δ GPT-4o – Human = 0.04, $p < 0.001$). This pattern suggests GPT-4o consistently outperforms humans across all levels, with the largest gap among lower-performing individuals, gradually diminishing toward higher-performing ones. The mean Δ GPT-4o – Human across all quantiles was 0.19.

This pattern held for the MRMET, where GPT-4o again showed the largest differences at lower quantiles (Δ GPT-4o – Human = 0.38 at the 0.03 quantile) (Table 3). As with the RMET, these differences decreased at higher quantiles but remained significant throughout ($p < 0.001$ for all quantiles), including at the 0.97 quantile (Δ GPT-4o – Human = 0.05, $p < 0.001$). This consistent pattern reinforces that GPT-4o's advantage is most notable among lower-performing individuals, with differences narrowing among high-performing participants. The mean Δ GPT-4o – Human across all quantiles was 0.20.

Figure 1 presents three complementary visualizations for both RMET and MRMET. Panels A and D show box plots of accuracy distributions for humans and GPT-4o, with quantile-based sampling applied to accommodate large sample sizes. Panels B and E display violin plots illustrating density estimates of accuracy, with vertical lines marking quantiles from 0.03 to 0.97 (thicker lines indicate medians). Panels C and F present shift-function plots comparing GPT-4o and human accuracy differences (Δ GPT-4o – Human) across quantiles, with 95% bootstrap confidence intervals shown as vertical lines. These visualizations confirm that GPT-4o's advantage is most pronounced among lower-performing individuals and gradually diminishes, though remains significant, at higher performance levels. To further establish this performance advantage across entire distributions, we next employed stochastic dominance tests.

**Figure 1. Performance and Distributional Difference between AI and Human in Emotional Recognition Tests**

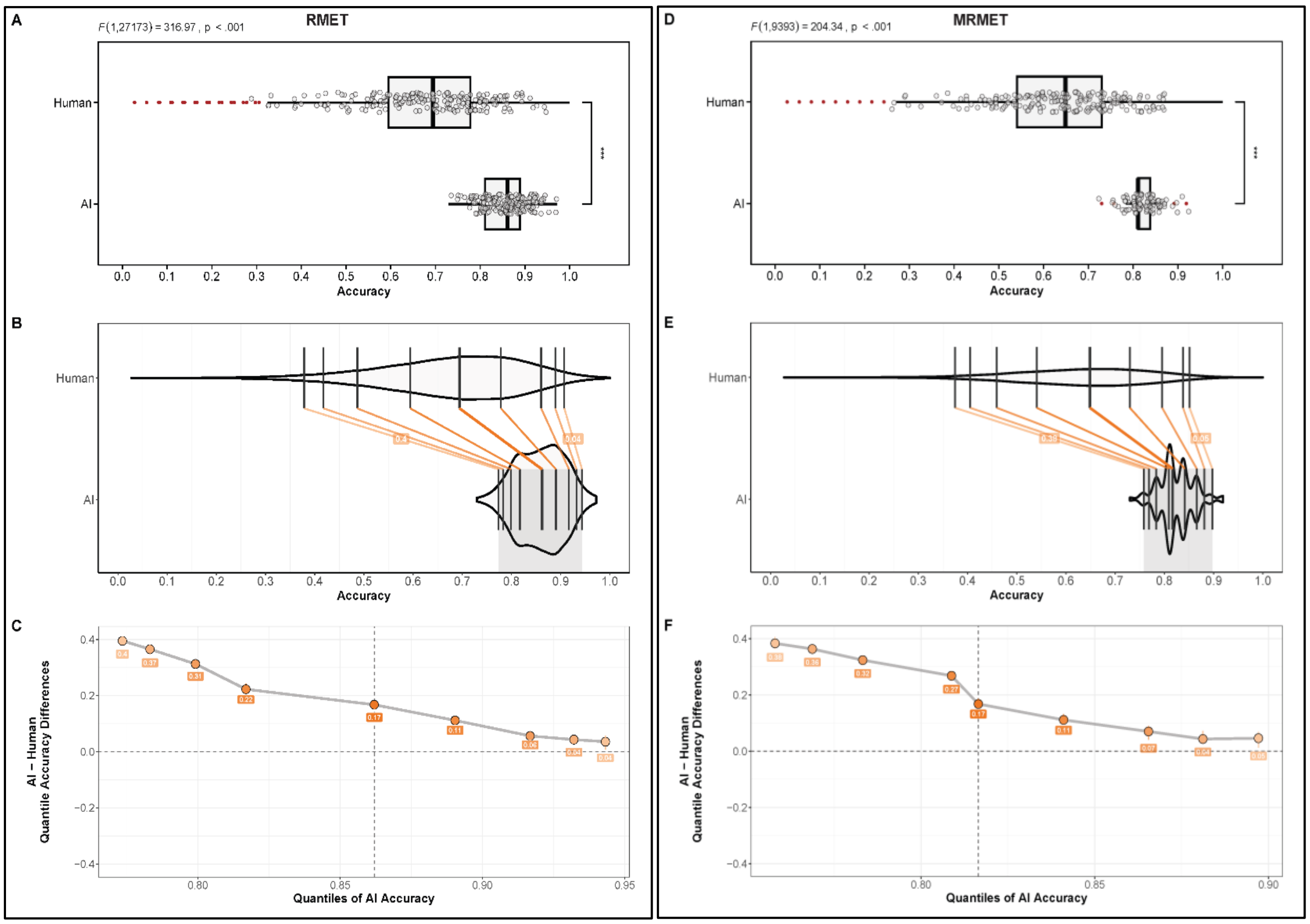

***Notes.*** **A & D.** Box plots show the mean of accuracy distributions for Humans and the GPT-4o (AI) condition on the RMET in panel A (N = 17,680 Humans) and MRMET in panel D (N = 9,295 Humans). For both datasets, the AI condition combines zero-shot and 10-shot prompting techniques, which showed no statistically significant differences and were thus pooled into a single AI group. Aggregated accuracy scores by participants were analyzed using ANOVA and reported in panel A for RMET and in panel D for MRET. To enhance visualization in the box plots, quantile-based sampling was applied to the Human datasets due to their large sample sizes. **B & E.** Violin plots illustrate the density estimates of accuracy for the Human and AI conditions on the RMET in panel B and MRMET in panel E. Vertical lines mark the quantiles from .03 to .97, with thicker lines indicating the medians. Effect sizes for specific quantiles were calculated using a bootstrapped shift function, with results for the .03 and .97 quantiles labeled in each respective figure. The violin plots highlight the performance distribution differences between Humans and AI for both datasets. **C & F.** Shift-function plots compare AI and Human accuracy (Δ AI – Human) across quantiles of the AI distribution for the RMET in panel C and MRMET in panel F. The vertical axes show the mean differences in accuracy for each quantile, with 95% bootstrap confidence intervals displayed as vertical lines. Confidence intervals that do not include zero indicate statistically significant differences. For both datasets, AI consistently outperforms Humans across all quantiles, but the effect size diminishes as quantiles increase. This pattern suggests a larger performance gap between lower-performing Humans and AI, while the gap between top-performing Humans and AI narrows.

## 3.3. AI Exhibits Stochastic Dominance Over Humans in Emotion Recognition

To further test whether GPT-4o's advantage holds across the entire performance spectrum, we conducted first-order stochastic dominance (FSD) and second-order stochastic dominance (SSD) tests comparing the distributions of GPT-4o and humans on both RMET and MRMET. Importantly, in both FSD and SSD tests, the null hypothesis states that one distribution dominates the other; thus, rejecting the null hypothesis indicates no evidence of dominance, while failing to reject it supports dominance. This interpretation reverses the typical statistical logic. In other words, to find evidence that one distribution stochastically dominates another, we seek p-values larger than 0.05 (i.e., we fail to reject the null), not smaller. Figure 2 visualizes these dominance results for both datasets.

For the RMET, the FSD test compared cumulative distribution functions (CDFs) of GPT-4o and humans. Testing whether humans dominated GPT-4o yielded Tobs = 10.29 ($p < 0.001$), rejecting human stochastic dominance. Testing whether GPT-4o dominated humans yielded Tobs = 0.01 ($p = 0.89$), implying GPT-4o's stochastic dominance (Supplementary Table S4). These results confirm GPT-4o demonstrated first-order stochastic dominance. In practical terms, at any accuracy level, GPT-4o has a higher probability of achieving that level or above than humans.

The SSD test further confirmed this pattern, rejecting human dominance over GPT-4o (Tobs = 184.77, $p < 0.001$) while implying GPT-4o dominance over humans (Tobs = 0.00, $p = 0.91$) (Supplementary Table S4). This finding indicates GPT-4o achieved higher overall accuracy while maintaining greater consistency across different test items.

The MRMET results paralleled those of the RMET, demonstrating performance differences favoring GPT-4o. The FSD test rejected human dominance (Tobs = 7.93, $p < 0.001$) while failing to reject GPT-4o dominance (Tobs = 0.01, $p = 0.86$) (Supplementary Table S4). The SSD test similarly rejected human dominance (Tobs = 69.94, $p < 0.001$) while failing to reject GPT-4o dominance (Tobs = 0.00, $p = 0.88$) (Supplementary Table S4). Thus, GPT-4o's better performance on the MRMET also spans the entire distribution.

These results collectively demonstrate that GPT-4o achieved both first- and second-order stochastic dominance on both RMET and MRMET. For any accuracy threshold, GPT-4o consistently has a higher probability of achieving that threshold or higher than humans. Second-order dominance further reveals both higher accuracy and greater stability across test items. Having established GPT-4o's individual superiority across the full performance spectrum, we now turn to our second research question: how does collective intelligence change these dynamics?

**Figure 2. First and Second-Order Stochastic Dominance**

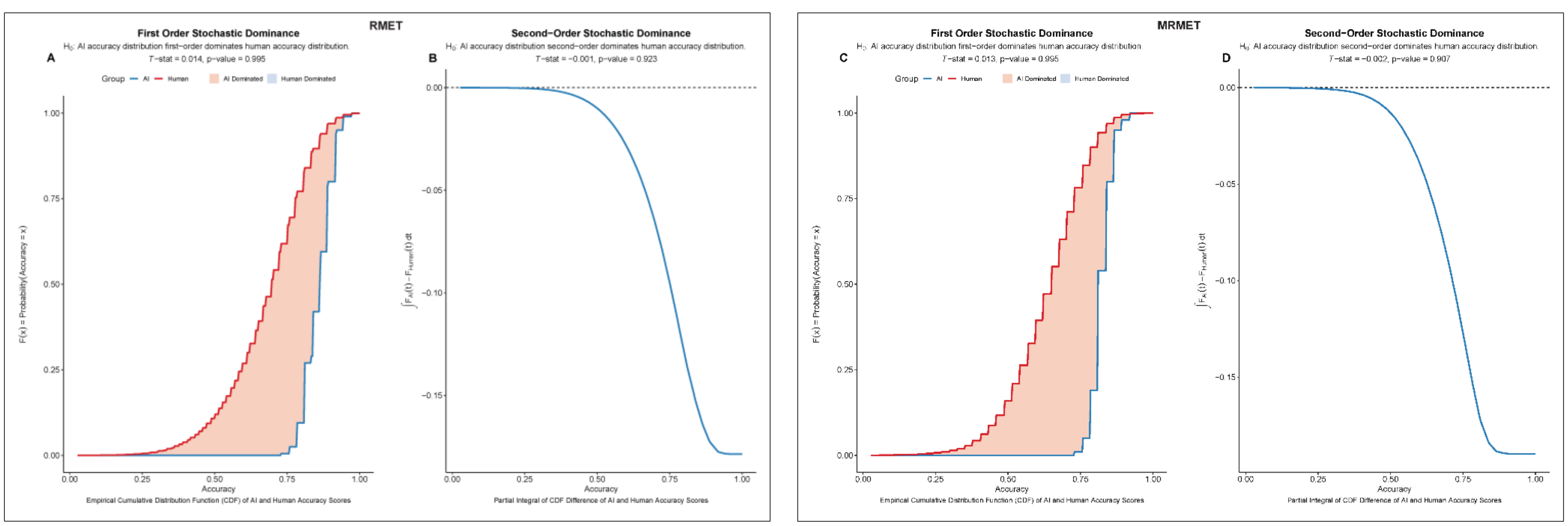

***Notes.*** We examined whether the AI accuracy distribution first-order (**Panels A** and **C**) and second-order (**Panels B** and **D**) stochastically dominates the Human accuracy distribution in both the RMET and MRET datasets. In all cases, the AI group combines zero-shot and n-shot trials due to the absence of significant differences between them. **Panel A** and **C** show the first-order dominance test results. We tested the null hypothesis that the AI accuracy distribution first-order dominates the Human accuracy distribution. This was done by comparing the empirical cumulative distribution functions (CDFs) of accuracy scores for

AI (blue) and Humans (red). For the RMET dataset (Panel A), the test yielded a T-statistic of 0.014 and a p-value of 0.995, while for the MRET dataset (Panel C), the T-statistic was 0.013 and the p-value was 0.995. In both cases, we failed to reject the null hypothesis, indicating that no statistically significant evidence was found to suggest that AI *violates* first-order dominance. These findings suggest that the data are consistent with AI *outperforming* Humans under the first-order dominance criterion. **Panels B** and **D** show the second-order dominance test results. We tested whether the AI accuracy distribution second-order dominates the Human accuracy distribution by evaluating cumulative differences in the CDFs (AI minus Human) across accuracy thresholds. For the RMET dataset (Panel B), the test yielded a T-statistic of -0.001 and a p-value of 0.923, while for the MRET dataset (Panel D), the T-statistic was -0.002 and the p-value was 0.907. Again, we failed to reject the null hypothesis in both datasets, indicating that no statistically significant evidence was found to suggest that AI *violates* second-order dominance. As with the first-order results, these findings imply that AI performance is consistent with *outperforming* Humans under the second-order dominance criterion.

### 3.4. Collective Intelligence: Aggregated Human Decisions Outperform GPT-4o in Emotion Recognition

To address our second research question, we examined whether GPT-4o can match or surpass the "wisdom of crowds" in humans when aggregating judgments through plurality voting. We formed GPT-4o crowds by aggregating responses from multiple independent runs and formed human crowds by randomly sampling participants, creating groups ranging from 5 to 90 members. Accuracy was assessed via generalized linear mixed-effects models (GLMMs), with fixed effects for condition (human as reference, GPT-4o, human + GPT-4o) and log-transformed group size, including their interaction. Random intercepts accounted for test items and iterations. Separate models were fitted for RMET and MRMET datasets.

**Table 4. Mixed-Effects Logistic Regression Results for Collective and Augmented Intelligence**

| Model Information | RMET | | | MRMET | | |
|---|---|---|---|---|---|---|
| **Fixed Effects** | **Estimate** | ***SE*** | ***p*-value** | **Estimate** | ***SE*** | ***p*-value** |
| (Intercept-Human Crowd) | 0.87 | 0.20 | < 0.001 | 1.20 | 0.30 | < 0.001 |
| GPT-4o Crowd | 1.92 | 0.07 | < 0.001 | 1.36 | 0.05 | < 0.001 |
| Augmented Intelligence | 0.04 | 0.06 | 0.51 | 0.11 | 0.05 | 0.030 |
| $Log_{10}$ Group Size | 3.99 | 0.06 | < 0.001 | 1.42 | 0.03 | < 0.001 |
| GPT-4o Crowd × $Log_{10}$ Group Size | -3.73 | 0.07 | < 0.001 | -1.47 | 0.04 | < 0.001 |
| Augmented Intelligence × $Log_{10}$ Group Size | 0.25 | 0.08 | 0.002 | 0.17 | 0.03 | < 0.001 |
| | | | | | | |
| **Random Effects** | | | | | | |
| Participant Variance | 0.26 | - | - | 0.19 | - | - |
| Participant *SD* | 0.51 | - | - | 0.44 | - | - |
| Item Variance | 1.37 | - | - | 3.27 | - | - |
| Item *SD* | 1.17 | - | - | 1.81 | - | - |
| | | | | | | |
| **Model Information** | | | | | | |
| Observations | 316,800 | - | - | 325,600 | - | - |
| Number of Participants | 6,600 | - | - | 6,600 | - | - |
| Number of Items | 36 | - | - | 37 | - | - |
| AIC | 64,491.600 | - | - | 158,731.800 | - | - |
| BIC | 64,576.900 | - | - | 158,817.400 | - | - |
| Log Likelihood | -32,237.900 | - | - | -79,357.900 | - | - |

**Note.** RMET = Reading the Mind in the Eyes Test; MRMET = Multiracial Reading the Mind in the Eyes Test. Both models were fit as generalized linear mixed models (GLMM) using the lme4 package in R, with a logit link function and binomial family distribution. Models were fit by maximum likelihood using adaptive Gauss–Hermite quadrature (nAGQ = 0) with crossed random intercepts for participant and item. Fixed-effects estimates are presented on the log-odds scale. *SE* refers to standard error. Human performance serves as the reference group for all comparisons (Intercept). The $Log_{10}$ Group Size variable represents the base-10 logarithm of aggregation size in **plurality voting**. Interaction terms (×) indicate how the effect of group size differs across conditions. Random effects are reported as variances and their standard deviations. AIC = Akaike Information Criterion; BIC = Bayesian Information Criterion. Both models showed successful convergence with no warnings or errors.

Table 4 presents the GLMM fixed effects for both tasks. On the RMET, the intercept (human baseline) was significant ($\beta = 0.87$, SE = 0.20, $p < 0.001$), with GPT-4o showing higher baseline log-odds ($\beta = 1.92$, SE = 0.07, $p < 0.001$) but a strongly negative interaction with log-group size ($\beta = -3.73$, SE = 0.07, $p < 0.001$), indicating minimal scaling benefits for AI crowds. The main effect of log-group size was positive and substantial for humans ($\beta = 3.99$, SE = 0.06, $p < 0.001$).

**Table 5. Marginal Probabilities and Contrast Results for Group Differences in Accuracy**

| Dataset | Condition/Contrast | Estimate | *SE* | 95% CI Lower | 95% CI Upper |
|---|---|---|---|---|---|
| **RMET** | | | | | |
| | **Marginal Probabilities** | | | | |
| | Collective Intelligence-Human | 0.998 | 0.000 | 0.998 | 0.999 |
| | Collective Intelligence-GPT-4o | 0.959 | 0.008 | 0.940 | 0.971 |
| | Augmented intelligence | 0.999 | 0.000 | 0.998 | 0.999 |
| | **Contrasts (Odds Ratios)** | | | | |
| | Human vs. GPT-4o | 26.015 | 1.710 | - | - |
| | Human vs. Augmented | 0.681 | 0.055 | - | - |
| | GPT-4o vs. Augmented | 0.026 | 0.002 | - | - |
| **MRMET** | | | | | |
| | **Marginal Probabilities** | | | | |
| | Collective Intelligence-Human | 0.960 | 0.012 | 0.930 | 0.977 |
| | Collective Intelligence-GPT-4o | 0.924 | 0.021 | 0.872 | 0.956 |
| | Augmented intelligence | 0.971 | 0.008 | 0.949 | 0.984 |
| | **Contrasts (Odds Ratios)** | | | | |
| | Human vs. GPT-4o | 1.954 | 0.043 | - | - |
| | Human vs. Augmented | 0.710 | 0.015 | - | - |
| | GPT-4o vs. Augmented | 0.363 | 0.007 | - | - |

**Note.** RMET = Reading the Mind in the Eyes Test; MRMET = Multiracial Reading the Mind in the Eyes Test. Estimates are derived from generalized linear mixed models. For Marginal Probabilities, estimates represent the predicted probability of correct responses for each condition, where *SE* = Standard Error and CI = Confidence Interval. All probabilities and confidence intervals are back-transformed from the logit scale. For Contrasts, odds ratios compare relative performance between conditions, where values greater than 1 indicate higher odds of correct responses for the first group in the contrast. All contrasts were significant at $p < 0.001$ (Bonferroni-adjusted for multiple comparisons). The Human + GPT-4o condition showed the highest predicted accuracy in both datasets, with more pronounced differences in the RMET dataset compared to MRMET.

Table 5 shows the marginal probabilities and odds ratio analysis. It revealed near-ceiling accuracy for human crowds (0.998, 95% CI [0.998, 0.999]) but lower for GPT-4o crowds (0.959, 95% CI [0.940, 0.971]). Odds ratios confirmed humans outperformed GPT-4o (OR = 26.01, $p < 0.001$). Slope analysis (Supplementary Table S7) underscored this pattern, showing that human crowds improved steeply with size (slope = 3.99, 95% CI [3.87, 4.10]), while GPT-4o showed limited gains (slope = 0.25, 95% CI [0.19, 0.32]).

On the MRMET, patterns were similar but with attenuated effects. The human baseline was significant ($\beta = 1.20$, SE = 0.30, $p < 0.001$), with GPT-4o again showing higher initial log-odds ($\beta = 1.36$, SE = 0.05, $p < 0.001$) but a negative interaction ($\beta = -1.47$, SE = 0.04, $p < 0.001$). Log-group size benefited humans ($\beta = 1.42$, SE = 0.03, $p < 0.001$; Table 4). Table 5 shows that marginal probabilities indicated human accuracy at 0.960 (95% CI [0.930, 0.977]), and GPT-4o at 0.924 (95% CI [0.872, 0.956]). Human crowds outperformed GPT-4o crowds (OR = 1.95, $p < 0.001$). Slope analysis (Supplementary Table S7) revealed human gains (1.42, 95% CI [1.37, 1.47]) but near-zero or slightly negative slopes for GPT-4o (-0.04, 95% CI [-0.09, 0.01]), suggesting aggregation may even reinforce errors in GPT-4o's predictions.

Figure 3 visualizes marginal trends by condition and group size for both tasks, illustrating how human crowds scale effectively while GPT-4o crowds plateau or decline slightly. These findings indicate that, despite GPT-4o's performance, its collective intelligence does not match human crowds, particularly as group size increases. The limited diversity in AI instances likely constrains the error-canceling benefits of aggregation, unlike the varied perspectives in human groups.

### 3.5. Augmented Intelligence: Combining Human and AI Leads to Optimal Performance in Emotion Recognition

Our third research question investigated whether augmented crowds, formed by combining independent human and GPT-4o votes at a 10:1 ratio, outperform human-only or GPT-4o-only crowds. Using the same plurality voting and generalized linear mixed-effects model (GLMM) framework, we included the human + GPT-4o condition in the models.

As shown in Table 4, on the RMET, the augmented intelligence (human + GPT-4o) condition had a non-significant baseline ($\beta = 0.04$, SE = 0.06, $p = 0.511$) but a positive interaction with log-group size ($\beta = 0.25$, SE = 0.08, $p = 0.002$), enhancing scaling beyond humans alone. In Table 5, marginal probabilities analysis confirmed the highest accuracy for augmented crowds (0.999, 95% CI [0.998, 0.999]), surpassing humans (OR = 1.47, $p < 0.001$, inverse of 0.68) and GPT-4o (OR = 38.17, $p < 0.001$, inverse of 0.03). Supplementary Table S7 shows the trend slope was steepest for augmented groups (4.23, 95% CI [4.13, 4.34]), compared to humans (3.99) and GPT-4o (0.25).

For the MRMET, augmented intelligence showed a positive baseline ($\beta = 0.11$, SE = 0.05, $p = 0.030$) and interaction ($\beta = 0.17$, SE = 0.03, $p < 0.001$) in Table 4. In Table 5, marginal probabilities yielded an accuracy of 0.971 (95% CI [0.949, 0.984]), outperforming humans (OR = 1.41, $p < 0.001$, inverse of 0.71) and GPT-4o (OR = 2.76, $p < 0.001$, inverse of 0.36). Supplementary Table S7 shows that trend slopes were highest for augmented (1.59, 95% CI [1.55, 1.64]), versus humans (1.42) and GPT-4o (-0.04).

Figure 3 further depicts these interactions, showing augmented crowds consistently achieving the greatest improvements with scale on both tasks. These results affirm RQ3, showing that augmented intelligence synergistically leverages human diversity and AI consistency, yielding superior emotion recognition without direct interaction.

**Figure 3. Artificial, Collective, Augmented Intelligence**

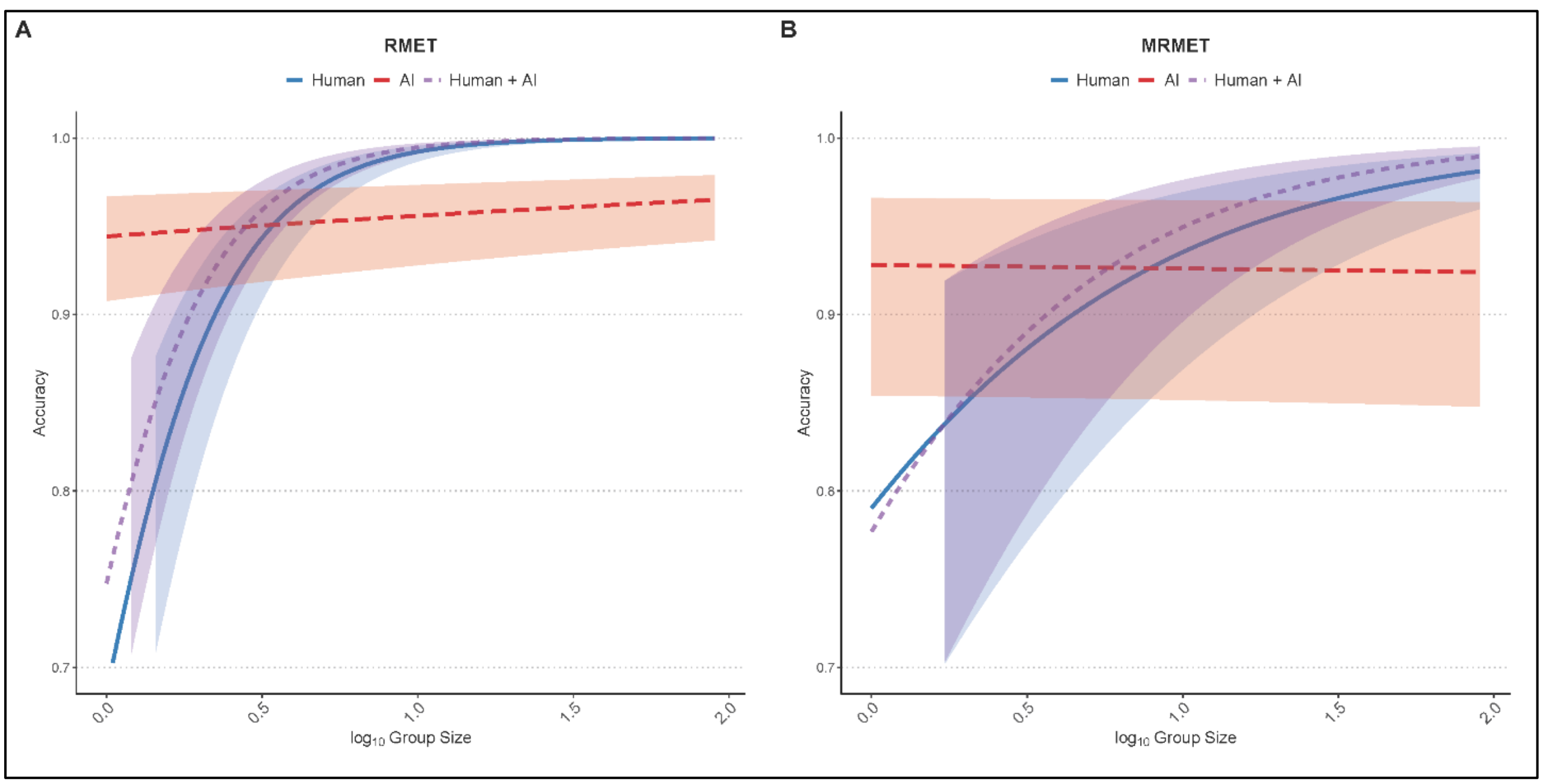

***Notes.*** Figure 3 illustrates how accuracy improves with larger group sizes (on a $\log_{10}$ scale) across three conditions—Humans (blue), AI (red), and Humans combined with AI (purple dashed)—for both the RMET dataset in Panel A (left) and the MRMET dataset in Panel B (right). The vertical axis shows average accuracy, and the shaded regions indicate approximate confidence intervals around each curve. In Panel A (RMET), the Human curve (blue) begins with relatively low accuracy for a single individual but rises sharply as group size increases, eventually reaching near-perfect performance. The AI curve (red) starts at a higher baseline but remains fairly steady, showing only modest improvement with group aggregation. The combined condition (purple dashed) benefits from both human collective intelligence and AI consistency, producing accuracy estimates that can exceed the levels observed for each group alone, particularly at larger group sizes. Panel B (MRMET) shows a similar pattern. Human accuracy (blue) again starts lower but improves substantially as the group gets larger, while the AI line (red) maintains a high yet fairly flat accuracy profile. When

humans and AI are combined (purple dashed), the resulting performance surpasses each condition individually for moderate to large group sizes.

# 4. Discussion

Our findings addressing the three research questions yield critical insights for understanding AI emotion recognition capabilities and limitations. First, GPT-4o exceeded human performance at all levels on RMET and MRMET, surpassing every performance group from the 3rd to 97th percentiles with first- and second-order stochastic dominance, confirming prior findings (Strachan, Albergo, et al., 2024; Tamkin et al., 2021). Second, aggregating judgments via plurality voting revealed a striking reversal: human crowds achieved dramatic accuracy improvements while GPT-4o crowds showed minimal gains. Third, augmented intelligence, which combines human cognitive diversity with GPT-4o's computational consistency, produced the highest accuracy, supporting ensemble-learning theory that uncorrelated classifiers surpass either alone (Domingos, 2000). These findings illuminate both AI capabilities and enduring human advantages. We discuss each finding's implications and limitations below.

The first finding of GPT-4o's individual superiority warrants careful interpretation. While one might interpret GPT-4o's performance advantage as demonstrating AI's superior emotion recognition, we believe it reveals as much about these standardized tests' limitations as it does about what AI can do. One such limitation is dataset contamination: GPT-4o might have been trained on these test items. Although GPT-4o's training corpus is proprietary, widely distributed RMET items likely appeared in pre-training data. The newer MRMET (released after GPT-4o's 2024 cut-off) had limited distribution, making contamination unlikely and suggesting genuine performance rather than memorization.

Furthermore, the magnitude of GPT-4o's advantage reveals important nuances. Accuracy differences between GPT-4o and humans varied systematically across ability levels. GPT-4o had large advantages over lower performers but modest advantages over higher performers. The advantage was 24% at the $10^{th}$ percentile and 6% at the $97^{th}$ for RMET; 27% and 5% for MRMET. This narrowing does not mean AI and human experts are equally skilled. Instead, it reflects a ceiling effect: when both score near 100%, tests cannot measure remaining differences. Expert humans may still possess finer abilities these tests cannot detect. Thus, our results speak to performance on standardized, static images, not naturalistic emotion recognition tasks where GPT-4o and human experts likely diverge significantly in their ability. These performance patterns have direct practical implications.

These nuanced performance differences provide preliminary guidance on the specific conditions under which GPT-4o can add incremental value alongside human judgment. In applied contexts, organizations and users might leverage AI for standardized, high-throughput screening in low-stakes conditions, reserving human expertise for ambiguous, context-dependent judgments that demand broader social understanding. However, this individual-level guidance changes dramatically when considering collective intelligence.

Our second finding addresses Research Question 2 about collective intelligence, revealing a striking reversal. When aggregating judgments via plurality voting, human crowds improved dramatically from 0.70 to perfect accuracy as groups grew, while GPT-4o crowds stagnated at 90–93%. The improvement rates, specifically 3.98 for humans versus 0.27 for GPT-4o, demonstrate clear collective human advantage. This reversal from individual AI superiority to collective human advantage demands explanation.

This collective advantage stems from a fundamental difference in how diversity emerges. Human crowds embody cognitive diversity from varied backgrounds and perspectives. Aggregating diverse perspectives cancels individual errors, creating stronger collective judgments. In contrast, GPT-4o showed no aggregation benefit. While temperature adjustments or alternative prompting could add diversity, these degrade real-world performance through reduced coherence (Peeperkorn et al., 2024). Remarkably, across all test items, the model's choices remained consistent regardless of sampling frequency. Each API call processes information through the same architecture and weights, yielding identical error patterns. Whereas humans generate varied predictions, GPT-4o produces uniform responses. Simply stated, repeated GPT-4o sampling resembles asking one person repeatedly, not polling a crowd. Consequently, aggregating GPT-4o responses reinforces rather than eliminates errors. Therefore, the wisdom of one AI differs fundamentally from the wisdom of many humans. This fundamental difference has critical implications for deployment decisions.

The wisdom-of-crowds advantage has critical practical implications for high-stakes applications. Human crowds remain indispensable for high-stakes emotion recognition tasks, despite requiring more resources and coordination. While it's efficient to deploy GPT-4o for large-scale emotion screening in low-stakes conditions, our results indicate we must prioritize human collective judgment when accuracy is vital, such as in clinical diagnoses, security screenings, or conflict mediations. The gap between human and AI collective judgments shows GPT-4o would misinterpret about 7% of emotional expressions. In these contexts, one misstep from AI could cause wrong diagnoses, miss security threats, or escalate conflicts. A single error could be costly. When deciding between humans or AI, organizations must consider whether they can accept any avoidable error. Given these constraints, a natural question emerges: can we achieve better performance by combining AI efficiency with human understanding?

This question leads to our third finding about augmented intelligence, which provides an affirmative answer. Augmented intelligence, created by combining human and AI judgments, achieved near-perfect accuracy (100% for RMET and 95% for MRMET). Its improvement rate as groups grew surpassed both human and AI conditions. While human-only crowds improved from 70% to 100% as group size increased, and GPT-4o crowds stagnated at 90–93%, augmented intelligence reached perfect accuracy using smaller group sizes. This superior performance emerges from complementary strengths.

This synergy emerges from a crucial complementarity between human and AI capabilities. Humans provide cognitive diversity, reducing unsystematic and uncorrelated errors. AI provides a consistent base level that corrects a group's typical mean accuracy. Keeping humans and GPT-4o independent in this process prevented humans from introducing errors and biases in the selection process and GPT-4o from causing overreliance. These two forms of intelligence, each bringing unique strengths and limitations, create through their augmentation emergent capabilities beyond what any single system can achieve. This aligns with distributed cognition theories (Hutchins, 1995), suggesting optimal emotion recognition requires collaboration between AI and human intelligence, distributed and combined independently. These findings point toward practical applications.

These findings point toward a new paradigm for intelligent systems development, one that favors augmentation and collaboration rather than replacement. Instead of developing fully autonomous AI that replaces human expertise, the target could be teams where AI is a critical member. Research shows that when humans and AI interact directly, performance degrades due to biases or cognitive shortcuts: cognitive offloading from over-relying on AI recommendations or algorithmic aversion bias from rejecting AI input. These established challenges in human-AI collaboration appear in medical diagnoses (Goh et al., 2024), education (Kim, 2024), and workplaces (Meng et al., 2025). Our research proposes an alternative approach: collecting independent judgments from human and AI and aggregating them through plurality voting proves more effective. Researchers might explore alternative ways to blend human and AI intelligence, including dynamic weighting schemes, hybrid committees with multiple AI models, and other strategies for augmented intelligence. This paradigm shift fundamentally reframes our approach to emotional AI.

Taken together, our three findings reveal a critical distinction in the current state of emotion recognition with AI. While GPT-4o's accuracy on standardized tests surpasses that of any individual, it cannot match the collective wisdom that emerges from diverse human crowds. The most promising path forward, however, is not a choice between human intuition and machine logic. Our findings conclusively show that an augmented intelligence strategy, which synergistically combines human and AI judgments, achieves a level of accuracy that neither can reach alone. This synergy capitalizes on both AI's consistent analytical rigor and humanity's collective emotional diversity. This understanding points to a clear directive for the future of affective computing. The objective should not be to build autonomous systems that supplant human insight, but to architect collaborative tools that amplify it. Ultimately, the development of truly effective and trustworthy emotional AI hinges on a paradigm shift, away from creating silicon minds that compete with us, and toward designing systems that enhance our collective human heart.